\documentclass[conference]{IEEEtran}
\IEEEoverridecommandlockouts
\usepackage{cite}
\usepackage{amsmath,amsfonts}
\usepackage{algorithmic}
\usepackage{graphicx}
\usepackage{textcomp}
\usepackage{xcolor}
\usepackage[ruled, vlined,linesnumbered]{algorithm2e}
\usepackage{epstopdf}
\usepackage{comment}
\usepackage{multirow}
\usepackage{url}
\usepackage{optidef}

\usepackage{amsthm}
\theoremstyle{definition}

\newcommand{\our}{\textit{SmartSplit}\xspace}
\newcommand{\comsnets}[1]{\textcolor{black}{#1}}

\def\BibTeX{{\rm B\kern-.05em{\sc i\kern-.025em b}\kern-.08em
    T\kern-.1667em\lower.7ex\hbox{E}\kern-.125emX}}
\begin{document}

\title{\our: Latency-Energy-Memory Optimisation for CNN Splitting on Smartphone Environment}

\author{\IEEEauthorblockN{Ishan Prakash}
\IEEEauthorblockA{\textit{Delhi Technological University}\\
New Delhi, India \\
ishan.0114@gmail.com}
\and
\IEEEauthorblockN{Aniruddh Bansal}
\IEEEauthorblockA{\textit{Delhi Technological University}\\
New Delhi, India \\
aniruddhbansal2000@gmail.com}
\and
\IEEEauthorblockN{Rohit Verma}
\IEEEauthorblockA{\textit{University of Cambridge}\\
Cambridge, UK \\
rv355@cam.ac.uk}
\and
\IEEEauthorblockN{Rajeev Shorey}
\IEEEauthorblockA{\textit{UQIDAR, IIT Delhi}\\
New Delhi, India \\
rshorey@iitd.ac.in}
\and
}

\maketitle

\begin{abstract}
Artificial Intelligence has now taken centre stage in the smartphone industry owing to the need of bringing all processing close to the user and addressing privacy concerns. Convolution Neural Networks (CNNs), which are used by several AI applications, are highly resource and computation intensive. Although new generation smartphones come with AI-enabled chips, minimal memory and energy utilisation is essential as many applications are run concurrently on a smartphone. In light of this, optimising the workload on the smartphone by offloading a part of the processing to a cloud server is an important direction of research. In this paper, we analyse the feasibility of splitting CNNs between smartphones and cloud server by formulating a multi-objective optimisation problem that optimises the end-to-end latency, memory utilisation, and energy consumption. We design \our, a Genetic Algorithm with decision analysis based approach to solve the optimisation problem. Our experiments run with multiple CNN models show that splitting a CNN between a smartphone and a cloud server is feasible. The proposed approach, \our fares better when compared to other state-of-the-art approaches.
\end{abstract}


\section{Introduction}~\label{introduction}
With the first set of AI applications launched on smartphones in 2017~\cite{micron}, the AI on smartphone industry has seen several improvements and use-cases. These use-cases, many of which utilise Convolution Neural Networks (CNNs)~\cite{lecun1999object}, include facial recognition, natural language translation, behaviour prediction, sensor data analysis for activity recognition, and many more~\cite{kong2016myshake,applefacrec}. Unlike other deep learning approaches that utilise matrix multiplication, CNNs utilise convolution in at least one of the network layers. Convolution is a resource and computation-intensive task. Seeing the importance of AI on smartphones the smartphone industry has been building specialised chips. The number of smartphones with specialised AI hardware has increased from $3\%$ in 2017 to $33\%$ in 2020 and is expected to grow further~\cite{counterpoint}. Huawei's Kirin processor~\cite{kirin} and Apple's Bionic AI Chip~\cite{appleaichip} are some of the popular examples of such specialised hardware.

A few crucial points need to be considered while running CNNs on the smartphone environment. First, although the new AI-enabled chips are suitable for running CNNs on smartphones, reducing resource and computation requirements is essential as the remaining resource or computation power could be utilised by other processes running simultaneously on the smartphone. Second, the AI-enabled chips are only available for the newer smartphones and this restricts the reach of several AI applications. Furthermore, a strategy that is likely work for earlier generation of smartphones would be useful for the newer ones to optimise already efficient AI-enabled chips.

Several studies have been done to expand the reach of AI on smartphones. One such direction of study develops compressed architecture of existing CNN models~\cite{sandler2018mobilenetv2, iandola2016squeezenet, yolov3}. However, these approaches are limited to the type of model they are compressing. Moreover, compressed architectures almost always compromise on accuracy. Another direction of research is to develop lightweight libraries that could be used to implement CNNs~\cite{tensorflowlite}. Some works prune the redundant and non-informative weights from the network architecture to develop models similar to the compressed models. However, both lightweight libraries and network pruning compromise on accuracy~\cite{khalil2021deep}. Another popular approach is splitting the CNN between the smartphone and a cloud server where a part of the processing is done on the cloud server. Such splitting could be done using N-step algorithms~\cite{mehta2020deepsplit} or using latency-based optimisation~\cite{tang2020joint}. These approaches have three major shortcomings when applied to smartphones. First, splitting-based approaches have been specifically designed for edge devices designed to run AI applications. The constraints that work in the context of edge devices would be different compared to smartphones. Smartphones are used for performing concurrent tasks that need to be taken into consideration. Second, these approaches optimise only a single system parameter, be it memory utilisation~\cite{mehta2020deepsplit} or latency~\cite{tang2020joint}. However, considering the multi-tasking capability of smartphones, we need to optimise both the parameters. Optimising latency is important; otherwise, the non-splitting approaches would be the better alternative. Furthermore, optimising memory utilisation is crucial as smartphone memory is used for additional parallel processes. Finally, none of the state-of-the-art works considers optimising the energy consumed when running computation-extensive applications.

In light of the limitations of the existing works, we formulate a multi-objective optimisation problem that optimises latency, memory utilisation, and energy consumption when a CNN is split between a smartphone and a cloud server. The key contributions in this paper are as follows;
\begin{itemize}
    \item Design a latency and energy model for a CNN being split between a smartphone and the cloud server ($\S~\ref{energymodel}$).
    \item Formulate a multi-objective optimisation problem that defines three objective functions to optimise latency, memory utilisation, and energy consumption along with the necessary constraints ($\S~\ref{problemdefinition}$).
    \item Design a genetic algorithm and decision analysis approach \our to solve the formulated multi-objective optimisation problem. ($\S~\ref{optimisationalgorithm}$).
    \item Design an Android Application that uses PyTorch for Android~\cite{pytorchandroid} for running the CNN models post splitting between the smartphone and the cloud server.
\end{itemize}
We evaluate our optimisation algorithm under varying conditions and compare it with different state-of-the-art approaches. Our results show that \our fares better than the existing approaches ($\S~\ref{evaluation}$). In the next section, we briefly describe the related literature in the area ($\S~\ref{relatedwork}$).
\section{Related Work}~\label{relatedwork}
Recent strides in Deep learning techniques, namely Convolution Neural Networks (CNN)~\cite{lecun1999object} have surpassed conventional computer vision techniques and have given superior accuracy for tasks ranging from image classification~\cite{lecun1999object, krizhevsky2012imagenet, sandler2018mobilenetv2, simonyan2014very, szegedy2017inception} to action recognition. Convolution Neural Networks are a class of Neural Networks using convolution operation in at least one of the layers. However, CNN-based techniques are computationally expensive as they have a large number of model parameters. The performance of CNN on mobile devices has been suboptimal compared to running CNN on superior computing hardware such as GPUs and Neuromorphic chips. 

Over the last few years, the computational power of mobile devices has increased drastically~\cite{huang2017deep}. This advancement prompts the development of techniques that will enable us to utilise the computing power by running CNNs on resource-constrained mobile devices. Several bottlenecks are encountered when running CNNs on mobile devices such as network delays, hardware limitations, and privacy concerns~\cite{ignatov2018ai}. There exist different techniques that can overcome the bottlenecks. Some techniques focus on CNN memory management; they address this problem by using feature map pruning and using certain encoding strategies to compress the neural network~\cite{khalil2021deep, han2015deep}. Memory reduction can also be achieved by designing efficient neural networks~\cite{teerapittayanon2016branchynet}. Another class of techniques focus on decreasing the complexity of the network~\cite{gamanayake2020cluster}. Tan et al.~\cite{tan2019mnasnet} design CNN models using automated search (AutoML). There are lightweight libraries that are optimised to run on mobile devices~\cite{matsubara2019distilled, tensorflowlite}. Yanai et. al.~\cite{yanai2016efficient} implement multi-threaded mobile applications by utilising either NEON SIMD~\cite{simd} instructions or the BLAS Library~\cite{nugteren2018clblast} for effective computation of convolution layers. 

The other class of techniques focuses on neural network splitting~\cite{latifi2016cnndroid}. In this technique, the neural network is divided between the mobile device and a cloud server. In tasks such as image recognition, there is no straightforward implementation. However, several strategies have been studied for optimal splitting of CNNs on edge devices~\cite{zhang2016adaptive,malekhosseini2020splitting,jin2019split,leroy2021optimal, mehta2020deepsplit}. Other techniques have studied the performance of the above approaches on cloud and mobile devices separately~\cite{guo2018cloud}. Eshratifar et al.~\cite{eshratifar2019bottlenet} introduce a splitting algorithm for reducing the size of the feature map before sending it to the server. The primary limitation of these techniques is that they depend on the architecture of the neural network and cannot be generalised for smartphones. 

\section{Latency and Energy Model}\label{energymodel}

In order to incorporate latency and energy consumption as objective functions in the optimisation problem, we first need to design a latency and an energy model. The latency model defines how much time is consumed performing different tasks in the splitting process. The energy model defines how much energy the smartphone consumes when splitting a CNN model between the smartphone and the cloud server. We first perform a pilot study to observe how latency and energy consumption vary when different CNN models are split between smartphones and the cloud server.

\subsection{Pilot Study}
We set up the experiment between two smartphones and a Windows 10 cloud server. The first smartphone is a Samsung Galaxy J6 and has 4 GB RAM, 64 GB internal storage, 3000 mAh battery. It runs on an Exynos 7 Octa 7870 processor. The second smartphone is a Redmi Note 8, has 4 GB RAM, 32 GB internal storage, 4000 mAh battery. It runs on a Qualcomm SDM665 Snapdragon 665 Octacore processor. Both devices run Android version 10 as the operating system. The cloud server runs a Windows 10 operating system with 8 GB RAM and a 1.6 GHz quad-core Intel i5 processor. All three devices are connected to one Wi-Fi network with bandwidth of 10 Mbps. We run our experiments with four pre-trained CNN models, Alexnet (21 layers)~\cite{krizhevsky2012imagenet}, VGG11 (29 layers), VGG13 (33 layers), and VGG16 (39 layers)~\cite{simonyan2014very}.

We design an Android application using PyTorch for Android~\cite{pytorchandroid} to run the CNN models on the smartphone. PyTorch for Android has the advantage that it does not prune or automatically quantise the CNN model, unlike other optimised deep learning libraries. PyTorch for Android ensures that there is no compromise on accuracy. The application stores an instance of a pre-trained model split at a certain layer (called the \textit{split index}) on the smartphone and loads it on execution. The application takes a set of images and the CNN model as input and performs an image classification task. As the experiment is run for all possible split indices for all models, we uploaded a different model instance for each run.

\subsubsection{Latency Analysis}
There are four contributing tasks involved in the end-to-end process of running a split CNN between the smartphone and the cloud server. These tasks are local computation of the CNN model on the smartphone (\textit{Client Latency}), sending the intermediate results to the cloud server (\textit{Upload Latency}), local computation of the remaining CNN layers at the cloud server (\textit{Cloud Server Latency}), and retrieving the final results from the cloud server (\textit{Download Latency}). The Android Application logs the timestamp when each task starts and the corresponding latency could be computed. However, we observe that the Download Latency is negligible and hence is not included in our results. Therefore, we plot the other three latency values and the corresponding total latency for every possible split index.

\begin{figure}[!ht]
  \centering
  \includegraphics[width=1\linewidth]{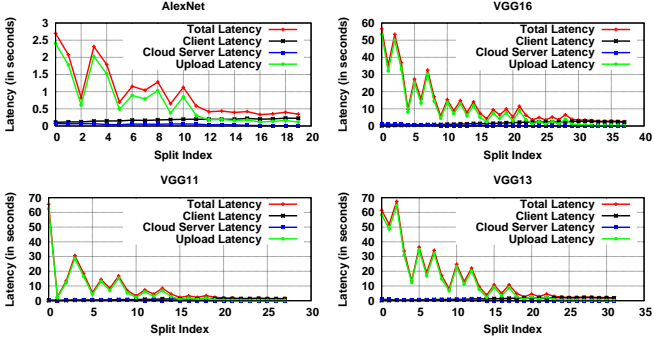}
  \caption{Latency vs CNN split index for Samsung J6.}
  \label{fig:j6_latency}
\end{figure}

\begin{figure}[!ht]
  \centering
  \includegraphics[width=1\linewidth]{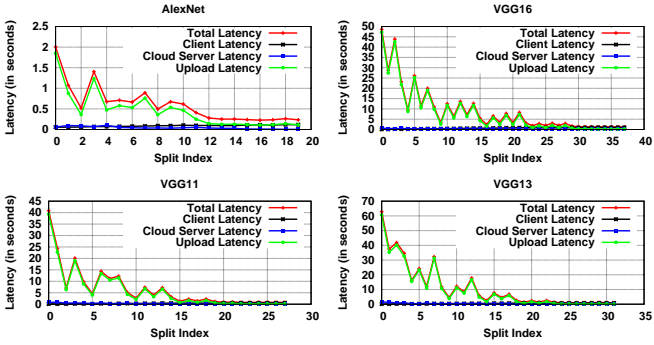}
  \caption{Latency vs CNN split index for Redmi Note 8.}
  \label{fig:mi_latency}
\end{figure}

Figure~\ref{fig:j6_latency} and Figure~\ref{fig:mi_latency} show the variation of the different latency values for all four CNN models when run on the two smartphones and split at all possible split indices. Both devices show similar trends for latency, with the Upload Latency being the primary contributing factor to the total latency. The Cloud Server Latency shows low variations because of the high resource availability at the cloud server. As expected, the Client Latency increases as more and more layers are processed on the smartphone.

\subsubsection{Energy Analysis}
The energy consumption calculations are done using the Android Battery Stats tool. For each run of the experiment, we retrieve the bug report file using the Android Debugger. The bug report file contains the voltage ($V$) and charge ($Q$) information at regular timestamps. With the voltage and charge values, we compute the energy consumed ($E$) as follows~\cite{oliveira2017study};
\begin{equation}
    \mathcal{E} = V * Q
\end{equation}

We consider the four contributing tasks. The cloud server computation would not contribute to energy consumption on the smartphone. We consider the other three tasks and calculate \textit{Client Energy}, \textit{Upload Energy}, and \textit{Download Energy}. Once the application run is complete, we compute the energy for the three tasks and add all three to get the \textit{Total Energy}. We plot the four energy values for all models when split at all possible split indices.

\begin{figure}[!ht]
  \centering
  \includegraphics[width=1\linewidth]{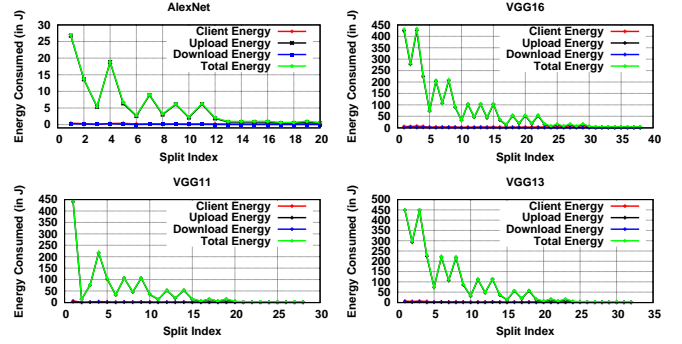}
  \caption{Energy consumption vs CNN split index for Samsung J6.}
  \label{fig:j6_energy}
\end{figure}

\begin{figure}[!ht]
  \centering
  \includegraphics[width=1\linewidth]{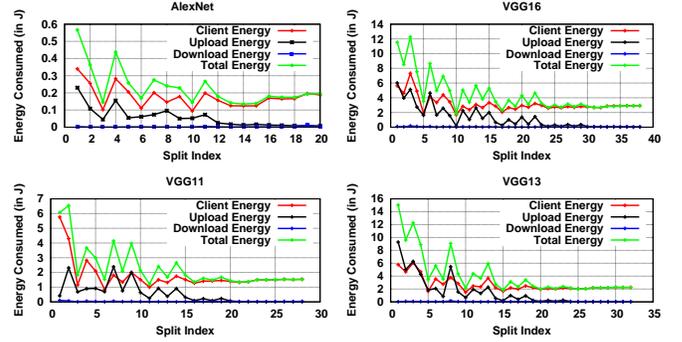}
  \caption{Energy consumption vs CNN split index for Redmi Note 8.}
  \label{fig:mi_energy}
\end{figure}

\begin{figure}[!ht]
  \centering
  \includegraphics[width=1\linewidth]{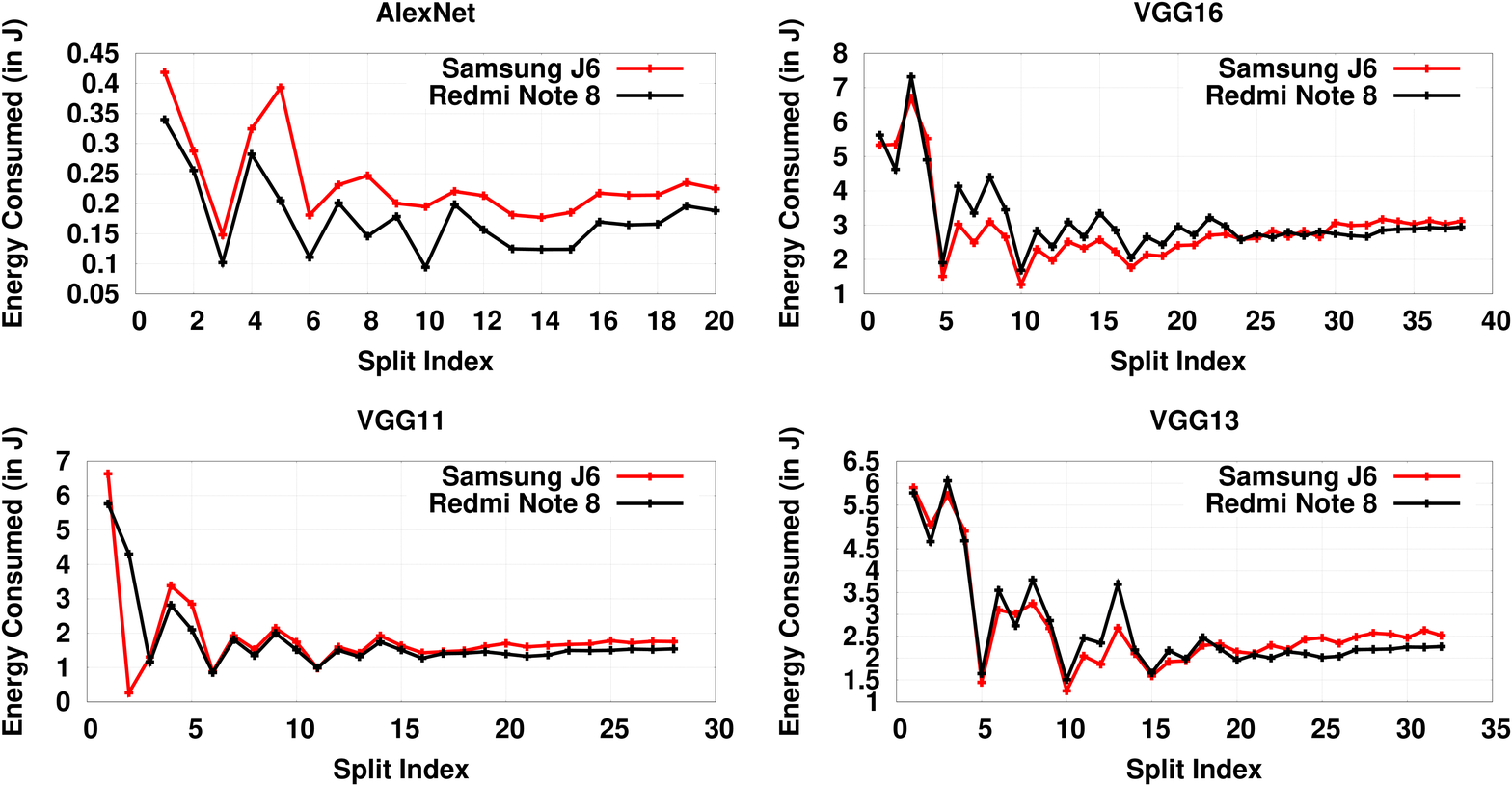}
  \caption{Client energy consumption for Samsung J6 and Redmi Note 8.}
  \label{fig:ce_compare}
\end{figure}

Figure~\ref{fig:j6_energy} and Figure~\ref{fig:mi_energy} show the variation of energy consumption at all split indices for all the four CNN models. We observe that the download energy is very low for all scenarios; however, contrasting results are obtained for the two smartphones for upload and client energy. The upload energy is the primary contributing factor for Samsung J6, while client energy is the primary contributing factor for Redmi Note 8. The contrasting result is due to the different WiFi standards followed by the two devices. Samsung J6 uses the WiFi 802.11 b/g/n standard, which consumes higher energy to perform the upload task. Redmi Note 8 uses the WiFi 802.11 ac standard, which is much more energy-optimised and hence consumes considerably less energy~\cite{sun2016experimental,noordbruis2020energy}. As can be seen in Figure~\ref{fig:ce_compare}, the client energy consumption remains almost similar for both the devices.

\subsection{Latency Model}
As discussed in the pilot study, three latency components contribute to the overall latency of running a CNN between the smartphone and the cloud server. We develop a model to compute the three components used to define the latency objective function.

\subsubsection{Client Latency ($\mathcal{T}_{client}$)}
Given that $l_1$ layers are computed on the smartphone, the client latency is given as;
\begin{equation}\label{eqn:clientlatency}
    \mathcal{T}_{client} = \frac{\mathcal{M}_{client}|l_1}{\mathcal{C}_{client} * \mathcal{S}_{client}}
\end{equation}
In Equation~\ref{eqn:clientlatency}, $\mathcal{M}_{client}|l_1$ is the total memory utilised when $l_1$ layers of the CNN model are run on the smartphone. The memory utilisation is computed using the tensor size, kernel size, and the type of CNN layer~\cite{params}. $\mathcal{C}_{client}$ is the number of cores, and $\mathcal{S}_{client}$ is the processor speed of the smartphone.

\subsubsection{Cloud Server Latency ($\mathcal{T}_{server}$)}
Given that $l_2$ layers are computed at the server we can calculate $\mathcal{T}_{server}$ as;
\begin{equation}\label{eqn:serverlatency}
    \mathcal{T}_{server} = \frac{\mathcal{M}_{server}|l_2}{\mathcal{C}_{server} * \mathcal{S}_{server}}
\end{equation}
In Equation~\ref{eqn:serverlatency}, $\mathcal{M}_{server}|l_2$ is the total memory utilised when $l_2$ layers of the CNN model are run on the server, $\mathcal{C}_{server}$ is the number of cores, and $\mathcal{S}_{server}$ is the processor speed of the cloud server.

\subsubsection{Upload Latency ($\mathcal{T}_{upload}$)}
The upload latency $\mathcal{T}_{upload}$ is dependent on the intermediate model size and the network bandwidth. It can be computed as;
\begin{equation}\label{eqn:uploadlatency}
    \mathcal{T}_{upload} = \frac{\mathcal{I}|l_1}{\mathcal{B}}
\end{equation}
In Equation~\ref{eqn:uploadlatency}, the intermediate model size given that $l_1$ layers are computed at the smartphone ($\mathcal{I}$) is dependent on the kernel size, stride, padding, and the type of layer~\cite{params}. The network bandwidth ($\mathcal{B}$) determines the transmission speed of the intermediate model to the cloud server.

The overall latency is represented as follows:

\begin{equation}
    \mathcal{T} = \frac{\mathcal{M}_{client}|l_1}{\mathcal{C}_{client} * \mathcal{S}_{client}} + \frac{\mathcal{I}|l_1}{\mathcal{B}} + \frac{\mathcal{M}_{server}|l_2}{\mathcal{C}_{server} * \mathcal{S}_{server}}
\end{equation}

\subsection{Energy Model}
The total energy consumed depends on three energy components, \textit{Client Energy}, \textit{Upload Energy}, and \textit{Download Energy}, as we have shown earlier in the pilot study. We compute the energy as a product of Power and Time. The time component is available using our latency model. The power consumption at the client and network transmission follow different models that we describe next.

\subsubsection{Client Energy ($\mathcal{E}_{client}$)}
The dynamic power consumption when executing an application on a smartphone is proportional to the number of cores ($\mathcal{C}_{client}$) and the operating frequency ($\nu$). The client power is expressed as follows~\cite{sundriyal2018modeling};
\begin{equation}\label{eqn:powerclient}
    P_{client} = k*\mathcal{C}_{client}*\nu^3
\end{equation}
where $k$ is an empirical constant. 

Using Equation~\ref{eqn:powerclient} and \ref{eqn:clientlatency}, we express the Client Energy $\mathcal{E}_{client}$ when computing $l_1$ layers at the smartphone as follows;
\begin{equation}\label{eqn:energyclient}
    \mathcal{E}_{client} = \left( k*\mathcal{C}_{client}*\nu^3 \right) * \left( \frac{\mathcal{M}_{client}|l_1}{\mathcal{C}_{client} * \mathcal{S}_{client}} \right)
\end{equation}
Using the results we obtain in the pilot study, we calculate the value of the constant $k$ as $1.172$.

\subsubsection{Upload Energy ($\mathcal{E}_{upload}$)}
We use the data transfer energy model given by Huang et al.~\cite{huang2012close} to compute the power consumption during model upload. Given that the upload throughput of the network is $\tau_u$, the upload power consumption is given as;
\begin{equation}\label{eqn:uploadpower}
    P_{upload} = \alpha_u * \tau_u + \beta_u
\end{equation}
where $\alpha_u$ and $\beta_u$ are empirical constants.
Using Equation~\ref{eqn:uploadpower} and ~\ref{eqn:uploadlatency}, we express the Upload Energy $\mathcal{E}_{upload}$ when sending the intermediate model after computing $l_1$ layers at the smartphone as follows;
\begin{equation}\label{eqn:uploadenergy}
    \mathcal{E}_{upload} = \left( \alpha_u * \tau_u + \beta_u \right) * \left( \frac{\mathcal{I}|l_1}{\mathcal{B}} \right)
\end{equation}

\subsubsection{Download Energy ($\mathcal{E}_{download}$)}
Similar to $\mathcal{E}_{upload}$, we use the data transfer energy model given by Huang et al.~\cite{huang2012close} to compute the power consumption during result download. Given that the download throughput of the network used is $\tau_d$, the download power consumption is given as;
\begin{equation}\label{eqn:downloadpower}
    P_{download} = \alpha_d * \tau_d + \beta_d
\end{equation}
where $\alpha_d$ and $\beta_d$ are empirical constants.
We need to define the time consumed when retrieving the result from the cloud server. We can express the download time as follows;
\begin{equation}\label{eqn:downloadlatency}
    \mathcal{T}_{download} = d/\mathcal{B}
\end{equation}
where $d$ is the download size. Using Equation~\ref{eqn:downloadpower} and~\ref{eqn:downloadlatency}, we express the Download Energy $\mathcal{E}_{download}$ when retrieving the result from the server as follows;
\begin{equation}\label{eqn:downloadenergy}
    \mathcal{E}_{download} = \left( \alpha_d * \tau_d + \beta_d \right) * \left( \frac{d}{\mathcal{B}} \right)
\end{equation}

With the three energy components defined, we model the overall energy consumption as follows;

\begin{equation}
    \begin{split}
        \mathcal{E} = \left( k*\mathcal{C}_{client}*\nu^3 \right) * \left( \frac{\mathcal{M}_{client}|l_1}{\mathcal{C}_{client} * \mathcal{S}_{client}} \right) \\ + \left( \alpha_u * \tau_u + \beta_u \right) * \left( \frac{\mathcal{I}|l_1}{\mathcal{B}} \right) \\ + \left( \alpha_d * \tau_d + \beta_d \right) * \left( \frac{d}{\mathcal{B}} \right)
    \end{split}
\end{equation}
We use the values of the constants as defined by Huang et al.~\cite{huang2012close}. These being;
\begin{center}
    $\alpha_u = 283.17$ mW/Mbps\\
    $\alpha_d = 137.01$ mW/Mbps\\
    $\beta_u = \beta_d = 132.86$ mW\\
\end{center}

With the latency and energy model in place, we now define the optimisation problem.
\section{Problem Definition}~\label{problemdefinition}
Recall that, pilot study shows that the latency and energy consumption varies with the layer at which the CNN model is split. Furthermore, we also observe that the variation in both latency and energy consumption is not monotonously increasing with split index. Moreover, since the smartphone memory is used for other applications running simultaneously, it is crucial that we also minimise the memory utilised by the application.

We define the three objective functions as follows;

\begin{equation}\label{eqn:f1}
    f_1 (l_1, l_2) = \frac{\mathcal{M}_{client}|l_1}{\mathcal{C}_{client} * \mathcal{S}_{client}} + \frac{\mathcal{I}|l_1}{\mathcal{B}} + \frac{\mathcal{M}_{server}|l_2}{\mathcal{C}_{server} * \mathcal{S}_{server}}
\end{equation}

\begin{equation}\label{eqn:f2}
    \begin{split}
        f_2 (l_1) = \left( k*\mathcal{C}_{client}*\nu^3 \right) * \left( \frac{\mathcal{M}_{client}|l_1}{\mathcal{C}_{client} * \mathcal{S}_{client}} \right) \\ + \left( \alpha_u * \tau_u + \beta_u \right) * \left( \frac{\mathcal{I}|l_1}{\mathcal{B}} \right) \\ + \left( \alpha_d * \tau_d + \beta_d \right) * \left( \frac{d}{\mathcal{B}} \right)
    \end{split}
\end{equation}

\begin{equation}\label{eqn:f3}
    f_3 (l_1) = \mathcal{M}_{client}|l_1
\end{equation}

Equation~\ref{eqn:f1} and Equation~\ref{eqn:f2} are directly derived from the Latency and Energy model respectively (Section~\ref{energymodel}), for variables $l_1$ and $l_2$, which are the number of CNN layers at the smartphone and on the cloud server. Equation~\ref{eqn:f3} represents the memory requirement when $l_1$ layers are computed at the smartphone. Based on these three equations, we formulate the optimisation problem as follows,

\begin{mini}|s|
{}{\mathbb{F} = (f_1, f_2, f_3)}
{}{}
\addConstraint{\mathcal{M}_{edge}|l_1 \leq \mathbb{M}}
\addConstraint{l_1 + l_2 = \mathbb{L}}{}
\addConstraint{1 \leq l_1 \leq \mathbb{L}}{}
\addConstraint{1 \leq l_2 \leq \mathbb{L}}{}
\addConstraint{\tau_u \leq \mathcal{B}}
\addConstraint{\tau_d \leq \mathcal{B}}
\label{eqn:optimizationproblem}
\end{mini}

Equation~\ref{eqn:optimizationproblem} gives the formulated optimisation problem that minimises $f_1$, $f_2$, and $f_3$ under the six constraints. The first constraint restricts the memory requirement from exceeding the total available memory at the smartphone ($\mathbb{M}$). The second constraint ensures that the number of layers on the smartphone and the cloud server add up to the total number of layers in the model ($\mathbb{L}$). The third and fourth constraints ensure at least one layer is computed at the smartphone and the cloud server. Finally, the last two constraints guarantee that the upload and download throughput is never above the available network bandwidth.
\section{The \our Optimisation Algorithm}\label{optimisationalgorithm}
In this section, we describe the \our algorithm that utilises a genetic algorithm along with decision analysis to solve the multi-objective optimisation problem formulated in the previous section. The results obtained from the pilot study (Section~\ref{energymodel}) show that an increasing trend in energy does not necessarily imply a strictly increasing trend in latency. Hence, a truly optimal solution cannot be obtained for the optimisation problem $\mathbb{F}$. Instead, Pareto-optimal~\cite{debreu1954valuation} results need to be calculated to obtain the Pareto set.

\begin{algorithm}
\SetAlgoLined
\KwResult{The Pareto Optimal Solution for $\mathbb{F}$}
    \tcp{NSGA-II to get the Pareto set of all Pareto optimal solutions}
    $\mathbb{O} \gets NSGA2(\mathbb{F})$ \tcp{The Pareto Set}
    \tcp{TOPSIS to get Pareto optimal solution from $\mathbb{O}$}
    $\mathcal{F} \gets \begin{bmatrix}
                            (f_1)_i & (f_2)_i & (f_3)_i
                        \end{bmatrix}_{3*n}$
                        
    $\mathcal{F}' \gets Column-Normalised(\mathcal{F})$
    
    $\mathcal{F}'' \gets \begin{bmatrix}
                            (f'_1)_i & (f'_2)_i & (f'_3)_i
                        \end{bmatrix}_{3*m}$ \tcp{The reduced matrix of size 3*m where the solutions match the constraints}
                        
    $(f'_j)_{ideal} \gets min_{i=1}^m(f'_j)_i\,\, where\, j \in [1,3]$
    
    $\mathcal{P} \gets [\sqrt{\sum_{j=1}^3 ((f'_j)_i - (f'_j)_{ideal})^2}]_{1*m}$
    
    $\mathbb{O}_{optimal} = \mathcal{F}_i$ corresponding to ${min(\mathcal{P})}$
    
 \caption{The \our Optimisation Algorithm}
 \label{algo:optimization}
\end{algorithm}

Algorithm~\ref{algo:optimization} shows the \our optimisation algorithm that has two parts. NSGA-II~\cite{deb2002fast} to get the Pareto set ($\mathbb{O}$), and TOPSIS~\cite{behzadian2012state} to get the best Pareto optimal solution ($\mathbb{O}_{optimal}$). The two parts of the algorithm are described next in the subsequent subsections.

\subsection{Pareto Set Computation using NSGA-II}
The Non-dominated Sorting Genetic Algorithm II~\cite{deb2002fast} has been designed to overcome the limitations of NSGA~\cite{srinivasan1994multi} and other state-of-the-art evolutionary strategies such as PAES~\cite{knowles1999pareto} and SPEA~\cite{zitzler1998evolutionary}. Employing elitism~\cite{du2018elitism}, a faster sorting algorithm, and ensuring diversity of Pareto-optimal solutions using crowding distance improves the convergence property of the algorithm. Furthermore, the NSGA-II provides solutions much closer to the Pareto front than other multi-objective optimisation approaches such as $\epsilon$-constrained optimisation~\cite{chankong2008multiobjective}, weighted sum~\cite{marler2010weighted}, or weighted metric methods~\cite{nour2009weighted}. We opt for NSGA-II to obtain the Pareto set in \our.

NSGA-II follows the standard genetic algorithm approach during execution; however, it modifies the mating and survival selection strategies. The individuals are first selected per the Pareto front, which leads to the fronts being split as not all individuals are able to survive~\cite{albadr2020genetic}. In these fronts that are created due to splitting, the solutions are selected based on \textit{crowding distance}. The crowding distance is the Manhattan distance in the objective space and is computed based on the density of solutions around each solution. NSGA-II also follows a binary tournament mating selection. Hence, every individual is compared based on their rank, followed by the crowding distance. Once all the Pareto optimal solutions are obtained to create a Pareto set ($\mathbb{O}$), we employ TOPSIS to get the best out of these solutions.

\subsection{Optimal Solution Estimation using TOPSIS}
In order to obtain the best solution from the Pareto set, we employ decision analysis based on TOPSIS~\cite{behzadian2012state}. Say the Pareto set ($\mathbb{O}$), of size $n$, contains the Pareto optimal solutions $\{ \mathbb{F}_1,\mathbb{F}_2,....,\mathbb{F}_n \}$. Then for every $i^{th}$ solution, the objective functions could be represented as; $\mathbb{F}_i = [(f_1)_i, (f_2)_i, (f_3)_i]$. For TOPSIS, we obtain the decision matrix for all solution in the pareto set as;
\[
\mathcal{F} = 
\begin{bmatrix}
    (f_1)_1 & (f_2)_1 & (f_3)_1\\
    (f_1)_2 & (f_2)_2 & (f_3)_2\\
    . & . & .\\
    . & . & .\\
    (f_1)_n & (f_2)_n & (f_3)_n\\
\end{bmatrix}
\]
The matrix elements in $\mathcal{F}$ are then column-wise normalised to obtain $\mathcal{F}'$. Next, we remove all the solutions that do not meet the constraints in $\mathbb{F}$ to get $\mathcal{F}''$ which has a reduced dimension of $3 * m$. Now, for each objective function, the ideal value is calculated as;
\begin{center}
    $(f'_j)_{ideal} \gets min_{i=1}^m(f'_j)_i$
\end{center}
where $j \in (1,3)$

The Euclidean distance between all elements of $\mathcal{F}$ and the ideal points is calculated to get the distance matrix as;
\[
\mathcal{P} = 
\begin{bmatrix}
    \sqrt{\sum_{j=1}^3 ((f'_j)_1 - (f'_j)_{ideal})^2}\\
    \sqrt{\sum_{j=1}^3 ((f'_j)_2 - (f'_j)_{ideal})^2}\\
    .\\
    .\\
    .\\
    \sqrt{\sum_{j=1}^3 ((f'_j)_m - (f'_j)_{ideal})^2}\\
\end{bmatrix}
\]
The solution corresponding to the minimum element in $\mathcal{F}$ is then selected as the Pareto optimal solution ($\mathbb{O}_{optimal}$) of the multi-objective optimisation problem $\mathbb{F}$.
\section{Performance Evaluation}~\label{evaluation}
In this section, we first describe the experimental setup followed by an evaluation of the \our algorithm. We then provide the comparison of \our with other state-of-the-art approaches.


\subsection{Experimental Setup}
We run our experiments on Samsung Galaxy J6 with 4 GB RAM, 64 GB internal storage, 3000 mAh battery, and an Exynos 7 Octa 7870 processor with an Android 10 operating system. The cloud server is a Windows 10 system with 8 GB RAM and a 1.6 GHz quad-core Intel i5 processor. We utilise a Wi-Fi connection with 10 Mbps bandwidth to connect both devices. For our experiments, we use five pre-trained CNN models: Alexnet (21 layers)~\cite{krizhevsky2012imagenet}, VGG11 (29 layers), VGG13 (33 layers), VGG16 (39 layers)~\cite{simonyan2014very}, and MobileNetV2~\cite{sandler2018mobilenetv2} (21 layers).

\subsection{Performance Evaluation of \our}

\begin{figure}[!ht]
  \centering
  \includegraphics[width=1\linewidth]{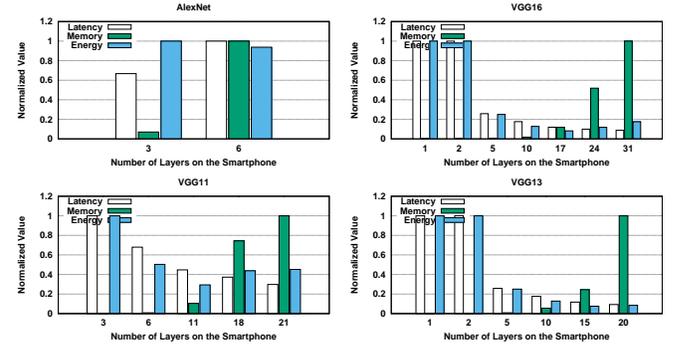}
  \caption{Values of Latency, Energy consumed, and Memory utilised for all the solutions in the Pareto set obtained using NSGA-II.}
  \label{fig:algonsga}
\end{figure}

\begin{table}[!ht]
\begin{tabular}{l|l|l|l|l|}
\cline{2-5}
                                                 & \textbf{AlexNet}       & \textbf{VGG11}          & \textbf{VGG13}          & \textbf{VGG16}          \\ \hline
\multicolumn{1}{|l|}{\textbf{Smartphone Layers}} & \multicolumn{1}{c|}{3} & \multicolumn{1}{c|}{11} & \multicolumn{1}{c|}{10} & \multicolumn{1}{c|}{10} \\ \hline
\end{tabular}
\caption{The optimal number of layers for the four models after applying TOPSIS on the Pareto set.}
\label{tab:algo_topsis}
\end{table}

We evaluate the \our algorithm to observe if the objective function values obtained based on the algorithm output are optimal or not. We test \our over AlexNet, VGG11, VGG13, and VGG16. In Figure~\ref{fig:algonsga} we plot the normalised latency, energy consumed, and memory utilised values for all solutions in the Pareto set obtained from NSGA-II. The corresponding best optimal solution obtained after performing TOPSIS is given in Table~\ref{tab:algo_topsis}. We observe that \our gives the optimal solution in all four cases. For example, when split at layer 3 in AlexNet, latency and memory utilisation are lowest, and the energy consumption is comparable. Similarly, for VGG11, when split at layer 11, memory utilisation and energy consumption are lower than other solutions in the Pareto set. Moreover, the latency is also comparable to the lowest value. In the case of VGG13 and VGG16, splitting at layer ten is close to the next best solutions, i.e., layer 15 for VGG13 and layer 17 for VGG16. However, in both CNN models, the memory utilisation is very low when splitting at layer 10, while the other two parameters are comparable to the next best solutions.

\subsection{Comparison with Competing Algorithms}
We compare \our with five competing algorithms that split the CNN between the smartphone and the cloud server based on other strategies.
\subsubsection{Latency-based Optimisation (LBO)} This class of optimisation algorithms only consider the best layer to split the CNN, such that the latency is minimum.

\subsubsection{Energy-based Optimisation (EBO)} Since energy-based optimisation algorithms are not available in the literature for smartphones; we design our own algorithm. In EBO, the optimal layer to split the CNN is the one that minimises the total energy consumed by running the CNN application on the smartphone. Effectively, EBO only has to solve $f_2$ of $\mathbb{F}$.

\subsubsection{CNN on Smartphone (COS)} This approach runs the complete CNN model on the smartphone.

\subsubsection{CNN on Cloud Server (COC)} This approach runs the complete CNN model on the server.

\subsubsection{Random Splitting (RS)} A random number is selected for each run, and the CNN is split at that random layer.

\begin{table}[!ht]
\centering
\begin{tabular}{|l|l|l|l|l|}
\hline
\textbf{Algorithm} & \textbf{AlexNet} & \textbf{VGG11} & \textbf{VGG13} & \textbf{VGG16} \\ \hline
\textbf{\our}       & 3                & 11             & 10             & 10             \\ \hline
\textbf{LBO}         & 3               & 21             & 20             & 25             \\ \hline
\textbf{EBO}         & 6               & 11             & 15             & 17             \\ \hline
\textbf{COS}         & 21               & 29             & 33             & 39             \\ \hline
\end{tabular}
\caption{\comsnets{Number of layers at smartphone for competing algorithms. COC has zero layers at the smartphone and RS generates random layers.}}
\label{tab:compete}
\end{table}

We evaluate all six competing algorithms over the four CNN models and compare the corresponding latency, energy consumption and memory utilisation values. in Table~\ref{tab:compete}, we provide the number of layers computed on the smartphone when using all the algorithms except RS and COC. COC has all the layers computed at the cloud server, and RS always gives a random value. We perform $100$ runs on Samsung Galaxy J6, and the average values are reported.

\begin{figure}[!ht]
  \centering
  \includegraphics[width=1\linewidth]{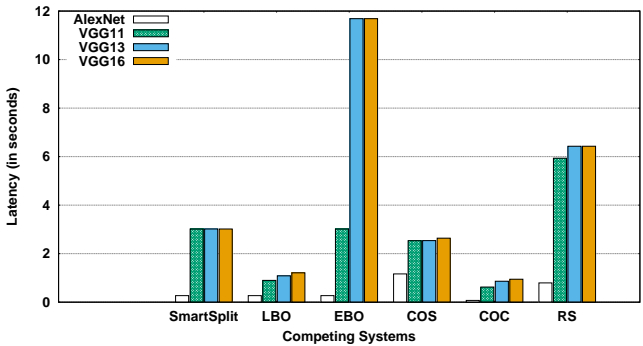}
  \caption{\textbf{Comparing latency values achieved when using different competing algorithms}}
  \label{fig:competing_latency}
\end{figure}

\begin{figure}[!ht]
  \centering
  \includegraphics[width=1\linewidth]{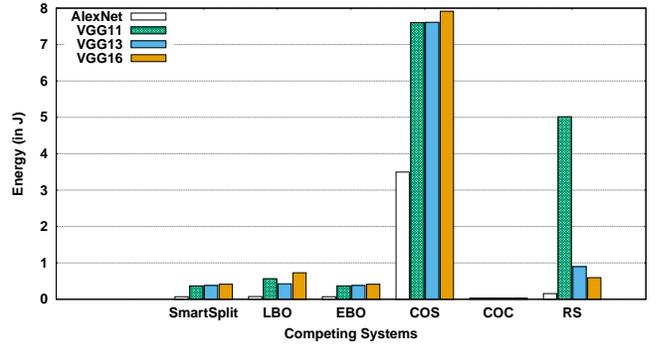}
  \caption{Comparing energy consumption values achieved when using different competing algorithms}
  \label{fig:competing_energy}
\end{figure}

\begin{figure}[!ht]
  \centering
  \includegraphics[width=1\linewidth]{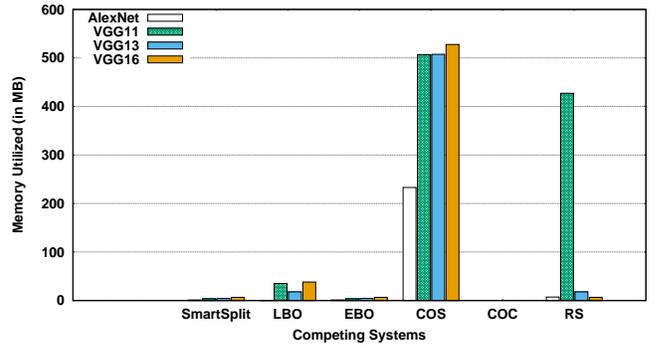}
  \caption{Comparing memory utilisation values achieved when using different competing algorithms}
  \label{fig:competing_memory}
\end{figure}

In Figure~\ref{fig:competing_latency}, Figure~\ref{fig:competing_energy}, and Figure~\ref{fig:competing_memory}, we plot the latency, energy consumption, and memory utilisation values for all competing algorithm respectively. Intuitively, the results obtained with RS is non-consistent owing to the random nature of the algorithm. COC, as expected, has the minimum values for latency and energy consumption and negligible memory requirement. However, it defeats the purpose of running AI on a smartphone environment. On the other hand, COS has the highest energy and memory consumption than the other competing algorithms because the smartphone has to perform all computation by itself. The COS algorithm's latency is comparatively low due to the absence of any interaction with the cloud server. Although the EBO algorithm has low energy consumption as expected and low memory consumption, it has high latency requirements. Moreover, the energy consumption values are also comparable to \our and LBO. LBO shows the closest results when compared to \our. However, \our has lower energy consumption and memory requirement than LBO and comparable latency values. The comparison results show that \our gives the best trade-off between latency, energy consumption and memory utilisation as compared to all the competing algorithms, thus making it a better alternative.

\subsection{Comparison with Smartphone Optimised Approach}
We compare \our with MobileNetv2~\cite{sandler2018mobilenetv2} which was explicitly designed to run on smartphones. MobileNetv2 has 21 layers and decreases the total number of model parameters using depth-wise separable convolutions~\cite{kaiser2017depthwise}, thus reducing the computational complexity and memory requirements. We compare MobileNetv2 with the four CNN models and the COS algorithm with VGG16.  We run our experiments for $100$ input images, and the reported results are averaged over $100$ runs.

\begin{figure}[!ht]
  \centering
  \includegraphics[width=1\linewidth]{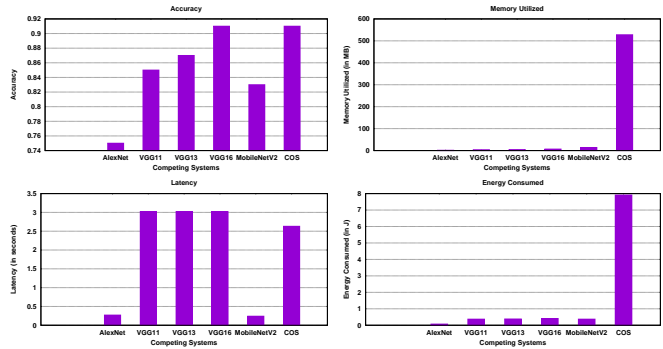}
  \caption{Comparison of \our with MobileNetV2 and COS}
  \label{fig:competing_models}
\end{figure}

In Figure~\ref{fig:competing_models}, we plot accuracy, latency, energy consumed and memory utilised values for the four CNN models that are split using \our, MobileNetv2, and COS. MobileNetV2 is a compressed model for smartphones and has almost $10\%$ less accuracy than VGG16 with \our. Furthermore, unlike COS and MobileNetV2, since not all layers of the CNN model is run on the smartphone, the memory utilised by the CNN models after splitting is lower than COS and MobileNetV2. Moreover, the energy consumed is similar to MobileNetV2 for the VGG models with split and lower for AlexNet with \our. MobileNetV2 does have lower latency. However, a latency difference of $\approx 2.7$ seconds is a small trade-off for improving accuracy by $10\%$ and a lower memory utilisation. We conclude from Figure~\ref{fig:competing_models} that splitting is a superior alternative for running CNNs on a smartphone environment.
\section{Conclusion}~\label{conclusion}
Smartphones are personal devices, and hence bringing AI-based applications to the smartphone has become crucial in recent times. Although the smartphone industry has been performing extensive research towards developing better hardware to run such applications, the earlier generation of smartphones without such hardware would miss out on the developments in the field. In this paper, we develop \our that makes resource-intensive AI-based applications accessible for most smartphones and ensures that resource utilisation is minimised, thus also favouring new smartphones with better hardware. We assert that running a CNN on a smartphone could be done more efficiently by splitting CNNs between the smartphone and a cloud server. We formulate this problem as a multi-objective optimisation problem that minimises the smartphone's end-to-end latency, memory utilisation, and energy consumed. We design a latency and energy model utilised to define the objective functions of the optimisation problem. In order to solve the multi-objective optimisation problem, we design \our, that uses NSGA-II to obtain a set of Pareto optimal solutions and then utilise TOPSIS to estimate the best Pareto optimal solution. Experiments on non-AI optimised smartphones show that \our gives optimal solution and performs better than other state-of-the-art approaches. The key takeaways from the paper are: (i) the upload energy is the primary factor affecting energy consumption in the earlier generation of smartphones, and hence network bandwidth is a crucial parameter to consider when splitting CNNs, (ii) \our is a better alternative compared to other splitting based approaches since it optimises three different objective functions, (iii) as compared to compressed CNN models, that compromise on accuracy, splitting the CNN between the smartphone and cloud server ensures higher accuracy and comparable latency along with low energy and memory consumption.

Some key aspects require further analysis in order to improve \our. First, generalising \our for other types of neural networks such as RNNs, LSTMs, or GANs would make more use-cases accessible on smartphones. Second, performing a splitting of CNNs in a centralised network topology. This direction of work would eliminate the single point of failure in \our. The third direction of work is to investigating how \our could be generalised such that smartphones are combined with other edge devices to create a heterogeneous edge ecosystem performing shared AI tasks such as localisation, smart home management, intruder detection, and several others. This work would involve investigating privacy concerns and fault mitigation.

\bibliographystyle{IEEEtran}
\bibliography{ref}

\end{document}